\documentclass[10pt,twocolumn,letterpaper]{article}

\usepackage[pagenumbers]{wacv}

\usepackage{graphicx}
\usepackage{amsmath}
\usepackage{amssymb}
\usepackage{booktabs}
\usepackage{algorithm}
\usepackage{algpseudocode}
\usepackage[pagebackref,breaklinks,colorlinks]{hyperref}
\usepackage[capitalize]{cleveref}

\crefname{section}{Sec.}{Secs.}
\Crefname{section}{Section}{Sections}
\Crefname{table}{Table}{Tables}
\crefname{table}{Tab.}{Tabs.}

\def\wacvPaperID{2722} 
\def\confName{WACV}
\def\confYear{2025}

\begin{document}


\title{Test-Time Adaptation of 3D Point Clouds via Denoising Diffusion Models}

\author{Hamidreza Dastmalchi$^{1}$, Aijun An$^{1}$, Ali Cheraghian$^{2,3}$, Shafin Rahman$^{4}$, Sameera Ramasinghe$^{5}$ \\
$^{1}$York University, Canada,
$^{2}$Data61-CSIRO, Australia,
$^{3}$Australian National University \\
$^{4}$ North South University, Bangladesh,
$^{5}$University of Adelaide, Australia \\
{\tt\small hrd@yorku.ca,~aan@yorku.ca,~ali.cheraghian@data61.csiro.au} \\ {\tt\small shafin.rahman@northsouth.edu,~sameera.ramasinghe@adelaide.edu.au}
}

\maketitle

\begin{abstract}
Test-time adaptation (TTA) of 3D point clouds is crucial for mitigating discrepancies between training and testing samples in real-world scenarios, particularly when handling corrupted point clouds. LiDAR data, for instance, can be affected by sensor failures or environmental factors, causing domain gaps. Adapting models to these distribution shifts online is crucial, as training for every possible variation is impractical.  Existing methods often focus on fine-tuning pre-trained models based on self-supervised learning or pseudo-labeling, which can lead to forgetting valuable source domain knowledge over time and reduce generalization on future tests. In this paper, we introduce a novel 3D test-time adaptation method, termed 3DD-TTA, which stands for 3D Denoising Diffusion Test-Time Adaptation. This method uses a diffusion strategy that adapts input point cloud samples to the source domain while keeping the source model parameters intact. The approach uses a Variational Autoencoder (VAE) to encode the corrupted point cloud into a shape latent and latent points. These latent points are corrupted with Gaussian noise and subjected to a denoising diffusion process. During this process, both the shape latent and latent points are updated to preserve fidelity, guiding the denoising toward generating consistent samples that align more closely with the source domain. We conduct extensive experiments on the ShapeNet dataset and investigate its generalizability on ModelNet40 and ScanObjectNN, achieving state-of-the-art results. The code has been released at \url{https://github.com/hamidreza-dastmalchi/3DD-TTA}.
\end{abstract}

\section{Introduction}
\label{sec:intro}

\begin{figure}[t]
    \centering
    \includegraphics[width=\columnwidth]{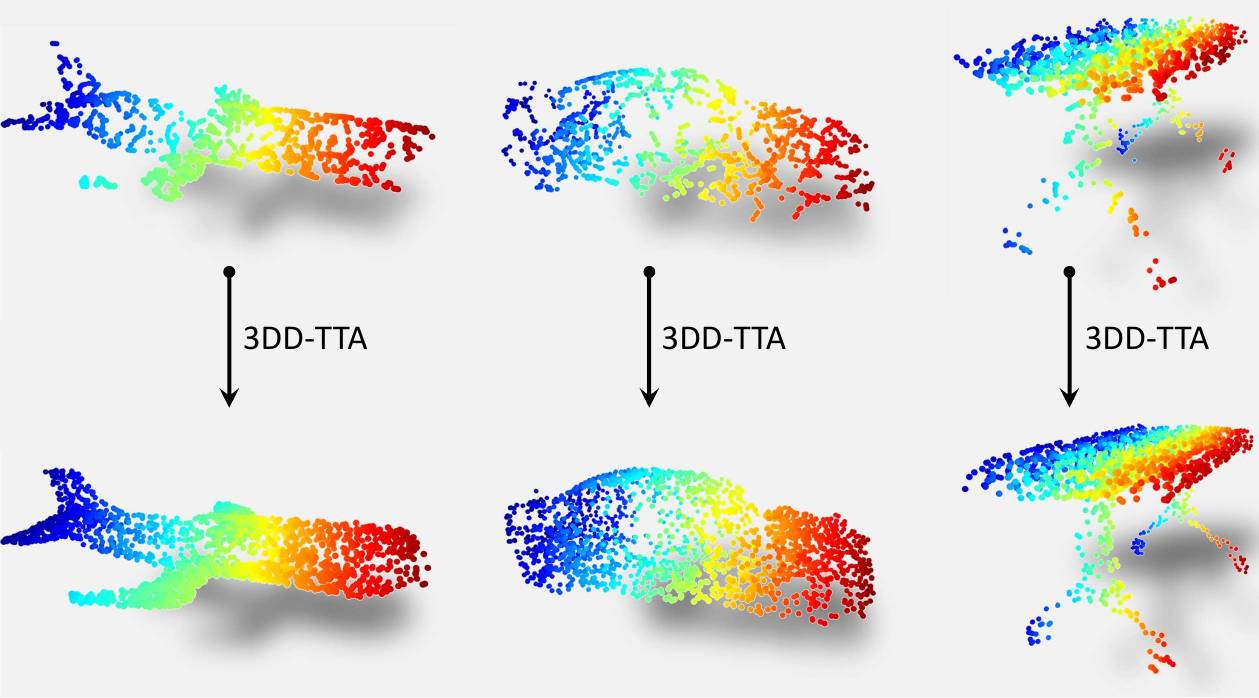}
    \caption{Reconstruction of corrupted point clouds using the proposed 3DD-TTA method.}
    \label{fig:denoising_process_samples}
\end{figure}

Point cloud processing has recently gained much attention and is critical in various computer vision applications \cite{qi2017pointnet, pointnet++, xiang2021walk, zhang2023starting, hong2023pointcam}. Although significant progress has been made in this field \cite{qi2017pointnet, pointnet++, xiang2021walk, zhao2021point}, much of the research has focused on controlled environments where the domain gap between training and testing samples is minimal. However, real-world scenarios often feature test samples that deviate from the training domain. For example, LiDAR point cloud data may be compromised by sensor failures or environmental factors, creating a domain gap that could lead to decreased performance. Moreover, distribution shifts in 3D data can vary greatly, rendering it infeasible to train networks for every possible test-time variation in point clouds. Therefore, it is crucial to efficiently adapt the models to unpredictable distribution shifts during the test stage in an online manner.

\begin{figure*}
    \centering
    \includegraphics[width=0.95\textwidth]{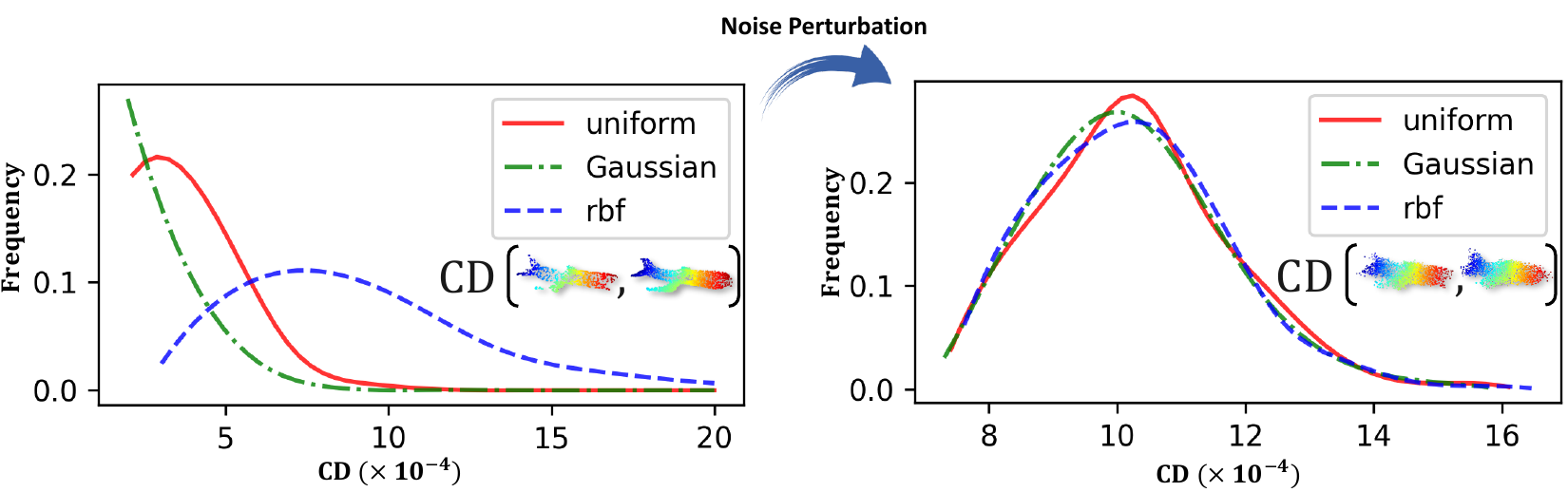}
    \caption{
     In the TTA setting, the source model encounters corrupted 3D point clouds with an unknown distribution shift, requiring adaptation without prior knowledge of the corruption type. Our 3DD-TTA approach adds Gaussian noise to the latent points (encoded by a pre-trained VAE) to reduce their dependence on the initial corruption. The distribution of the Chamfer Distance (CD) between original and corrupted point clouds—corrupted with uniform noise, Gaussian noise, and RBF \cite{rbf}—from ShapeNet dataset \cite{shapenet} is shown before \textbf{(Left)} and after \textbf{(Right)} Gaussian perturbation. After perturbation, the CD distributions for all corruption types overlap, demonstrating \emph{corruption independence}.
    }
    \label{fig:ood-independent}
\end{figure*}

Current approaches for test-time adaptation often involve fine-tuning the parameters of a pre-trained source model during inference. Some methods~\cite{Wang_2024_WACV, dobler2022robust, tent} utilize pseudo-labeling, adjusting the source model using pseudo-labels generated from test samples. Other techniques~\cite{liu2021ttt++, fan2022self, mirza2023mate} apply self-supervised learning to update the model for new test instances. However, both strategies face a common challenge: they may initially perform well but risk forgetting valuable source domain knowledge over time. It promotes error accumulation during the long TTA process. This can diminish the model's generalization ability, leading to subpar performance on future test samples. 

Unlike previous TTA methods \cite{Wang_2024_WACV, dobler2022robust, tent, liu2021ttt++, fan2022self, mirza2023mate}, which focus on updating the source model parameters and risk inducing forgetting, \cite{gao2023back, tsai2024gda, song2023target} introduced an alternative approach: adapting input test images to the source domain using deep generative models, such as denoising diffusion models (DDM) \cite{ddpm, ddm_thermodynamics}. While this approach has proven effective for 2D images \cite{gao2023back, tsai2024gda, song2023target, prabhudesai2023diffusion}, applying it to 3D point clouds presents a far greater challenge due to the unstructured nature of point clouds and the inherent difficulty in preserving their fidelity during the denoising process. Despite the growing success of diffusion models for 2D TTA, no existing method extends this powerful framework to 3D point clouds—until now.


To this end, we introduce a novel, training-free test-time adaptation method called 3D Denoising Diffusion Test-Time Adaptation (3DD-TTA). By training-free, we mean that the method adapts corrupted point clouds without modifying or fine-tuning the network parameters during the test phase. Specifically, our method employs a latent diffusion model, initially trained on the source domain, to adapt corrupted point clouds back to the source domain (see Figure \ref{fig:denoising_process_samples}). To enhance time efficiency, we reduced the computational overhead of our adaptation method by using a Denoising Diffusion Implicit Model (DDIM) \cite{ddim}, which skips denoising steps without major performance loss. Since point clouds typically lack high-frequency content, fewer denoising steps are sufficient to maintain performance. Before applying the reverse diffusion process in the latent space, we add Gaussian noise to the latent points (obtained from a VAE network) to reduce their dependence on variations in data distribution (see Figure \ref{fig:ood-independent}). To guide the denoising process and ensure it does not generate an inconsistent point cloud, we employ a modified Chamfer distance, termed Selective Chamfer Distance (SCD). The SCD measures the discrepancy between the original and predicted latent points at each time step, with updates made based on the gradient of this distance. This method improves fidelity while accommodating variations in the latent points.  Ultimately, we obtain a refined point cloud that is better aligned with the source domain, preserving the essential characteristics necessary for accurate classification. This alignment allows the source classifier to leverage its pre-trained knowledge more effectively, resulting in improved accuracy and reliability during the classification process.

In summary, our contributions are:
\textbf{{(1)}}  The paper introduces a novel, training-free method called 3D Denoising Diffusion Test-Time Adaptation (3DD-TTA), which  adapts corrupted point clouds rather than altering network parameters. To the best of our knowledge, we are the first to employ denoising diffusion models for TTA of 3D point clouds.
\textbf{{(2)}} We generate corruption-irrelevant latent points by introducing Gaussian noise to perturb latent points, which serve as inputs to a pre-trained latent diffusion model. This model then denoises the perturbed latent points iteratively through the reverse process, effectively restoring their alignment with the source domain characteristics. 
\textbf{{(3)}} We introduced a modified Chamfer distance, named Selective Chamfer Distance (SCD), to increase the fidelity during the reverse diffusion process. The SCD allows for targeted updates to the latent points and shape latent, ensuring that fidelity is maintained while also permitting necessary variations. This mechanism improves the model’s adaptability in handling corrupted data, ensuring more accurate reconstructions. 
\textbf{{(4)}} We conduct extensive experiments validating the approach on ShapeNet \cite{shapenet} ModelNet40 \cite{modelnet40}, and ScanObjectNN \cite{scanobjectnn} achieving new state-of-the-art results.

\section{Related Work}
\label{sec:related_work}

\noindent \textbf{Point Cloud Processing:}
PointNet~\cite{pointnet2017}, a pioneering work by Qi~\etal{}, marked the beginning of using multi-layer perceptron (MLP) networks for processing 3D point clouds. However, this initial method did not capture the intricate local structures inherent in the input data. To overcome this limitation, subsequent advancements like PointNet++~\cite{pointnet++2017} were introduced, employing hierarchical feature extraction to better capture local information. Building on this, various studies~\cite{RS-CNN2019,spidercnn2018,SPHNet2019,SFCNN2019,pointconv2019,pointcnn2018, chowdhury2022few, ahmadi2024accv, C3PR_ECCV} have proposed convolutional strategies to extract local information. 
Concurrently, other research efforts~\cite{dgcnn2019,LocalSpecGCN2018,PointGCN2018} have represented point clouds as graph vertices, enabling feature extraction in spatial or spectral domains. For instance, DGCNN~\cite{dgcnn2019} constructs graphs in feature space and dynamically updates them using MLPs for each edge. In contrast, PointGCN~\cite{PointGCN2018} employs k-nearest neighbors to generate graphs directly from point clouds, efficiently capturing local information.





\vspace{0.2cm}
\noindent \textbf{Diffusion Models:} 
Denoising Diffusion Models (DDMs) \cite{ddpm, ddm_thermodynamics} are a class of generative models that synthesize data by systematically reversing the forward diffusion process. This reverse process effectively transforms noisy inputs into high-quality outputs that align closely with the original data distribution. DDMs are especially noted for their ability to generate intricate images and complex data forms~\cite{glide, dalle2, imagen, latent-diffusion-model, sdxl-turbo}. Applying generative models to point clouds provides a powerful framework for unsupervised representation learning, effectively capturing data distribution. However, the development of diffusion models for 3D point clouds lags behind their image counterparts, largely due to the challenges posed by the irregular sampling patterns of point clouds in 3D space \cite{3d-diffusion}. However, a few diffusion models have been introduced for 3D point clouds \cite{3d-diffusion-gradient, point-voxel-3d-diffusion, 3d-diffusion, lion}. The LION model \cite{lion} distinguishes itself by applying a diffusion model to a hierarchical latent space, combining the global shape latent and local latent points to enhance expressivity and performance. We leveraged this latent diffusion model for the TTA task, utilizing its hierarchical framework to effectively adapt 3D point clouds.

\vspace{0.2cm}
\noindent \textbf{Test-Time Adaptation:} Test-time adaptation (TTA) adapts a source-trained model to a new target domain during inference without access to the source data. Common strategies include fine-tuning with an unsupervised loss on unlabeled target samples, generally designed for image domain adaptation. TTT~\cite{sun2020test} uses a self-supervised proxy task, while TENT~\cite{wang2021tent} updates batch normalization parameters by minimizing model response entropy. Similarly, SHOT~\cite{liang2020we} minimizes prediction entropy. TTT++~\cite{liu2021ttt++} adds a self-supervised branch with contrastive learning. MM-TTA~\cite{shin2022mm} merges predictions from multiple modalities using intra- and inter-modules for pseudo-labels. DSS~\cite{Wang_2024_WACV} addresses noisy pseudo-labels with dynamic thresholding and learning processes. For 3D point clouds, several TTA methods are specifically designed. MATE~\cite{mirza2023mate} employs masked autoencoding for robustness to distribution shifts in 3D point clouds. Hatem \textit{et al.}~\cite{hatem2023test} uses meta-learning for point cloud upsampling. Point-TTA~\cite{hatem2023point} adapts model parameters instance-specifically during inference for point cloud registration. In contrast to the fine-tuning methods, the authors in \cite{gao2023back} proposed a data-driven approach for TTA of 2D images using DDMs. Similarly, \cite{tsai2024gda, song2023target} employed diffusion models with structural guidance to enhance generalization for 2D image adaptation. In this work, we introduce a novel approach for TTA in 3D point clouds, leveraging a latent diffusion model with 3D shape guidance to address different distribution shifts.

\section{Method}
\label{sec:method}

\textbf{Problem Formulation:} Consider corrupted point clouds, each containing \(n\) points \(\tilde{\textbf{x}} \in \mathbb{R}^{n \times 3}\), belonging to a target domain \(\mathcal{Q}_t\). We have a classifier \(p_c\) and a  diffusion model \(\boldsymbol{\epsilon}_{\theta}\), both trained on original point clouds \(\textbf{x} \in \mathbb{R}^{n \times 3}\) from a source domain \(\mathcal{Q}_s\), where a domain gap exists between \(\mathcal{Q}_t\) and \(\mathcal{Q}_s\). The goal during test-time adaptation is to improve classifier accuracy on the corrupted data by adapting it back to \(\mathcal{Q}_s\) using \(\boldsymbol{\epsilon}_{\theta}\), without accessing the source data or modifying the classifier \(p_c\).




\subsection{Preliminaries}

\textbf{Denoising Diffusion Models:} Denoising diffusion models (DDMs) \cite{ddpm, ddm_thermodynamics} are generative models that learn the data distribution \( q(\textbf{x}) \) by reversing a forward diffusion process. In this process, Gaussian noise is progressively added to the data in a Markovian manner, and the model is trained to recover the original data from this noisy sequence.
 The forward diffusion process is defined as:
\begin{equation}
\begin{aligned}
q(\textbf{x}_{1:T} | \textbf{x}_0) &:= \prod_{t=1}^{T} q(\textbf{x}_t | \textbf{x}_{t-1}), \\
q(\textbf{x}_t | \textbf{x}_{t-1}) &:= \mathcal{N}\left(\textbf{x}_t; \sqrt{1 - \beta_t}\textbf{x}_{t-1}, \beta_t \textbf{I}\right)
\end{aligned}
\end{equation}

Here, \( T \) denotes the number of diffusion steps. \( q(\textbf{x}_t | \textbf{x}_{t-1}) \) represents the Gaussian transition distribution, and \( \beta_1, ..., \beta_T \) constitute a variance schedule. The choice of \( \beta_t \) aims to ensure that the chain approximately converges to a standard Gaussian distribution after \( T \) steps, i.e., \( q(\textbf{x}_T) \approx \mathcal{N}(\textbf{0}, \textbf{I}) \). According to the Markov chain property, for any intermediate timestep \( t \), we have:

\begin{equation}
\begin{aligned}
q(\textbf{x}_t | \textbf{x}_0) &= \mathcal{N}\left(\textbf{x}_t; \sqrt{\bar{\alpha}_t}\textbf{x}_0, (1 - \bar{\alpha}_t)\textbf{I}\right), \\
\textbf{x}_t &= \sqrt{\bar{\alpha}_t}\textbf{x}_0 + \sqrt{1 - \bar{\alpha}_t}\boldsymbol{\epsilon}
\end{aligned}
\label{eq:direct-forward-diffusion}
\end{equation}

where \( \bar{\alpha}_t := \prod_{s=1}^{t} (1 - \beta_s) \) and \( \boldsymbol{\epsilon }\) is drawn from a standard Gaussian distribution. Similarly, the generative process is modeled as a Gaussian transition with a learned mean \( \boldsymbol{\mu}_\theta \):
\begin{equation}
p_\theta(\textbf{x}_{t-1} | \textbf{x}_t) = \mathcal{N}\left(\textbf{x}_{t-1}; \boldsymbol{\mu}_\theta(\textbf{x}_t, t), \sigma_t^2 \textbf{I}\right)    
\end{equation}

Here, \( \sigma_t \) is typically a predefined constant associated with the variance schedule, and \( \boldsymbol{\mu}_\theta(\textbf{x}_t, t) \) is parameterized by a denoising U-Net, denoted by \( \epsilon_\theta(\textbf{x}_t, t) \), with the following equivalence \cite{ddpm}:
\begin{equation}
\boldsymbol{\mu}_\theta(\textbf{x}_t, t) = \frac{1}{\sqrt{\alpha_t}} \left(\textbf{x}_t -  \frac{1 - \alpha_t}{\sqrt{1 - \bar{\alpha}_t}} \epsilon_\theta(\textbf{x}_t, t)\right)    
\end{equation}
The denoising network is trained by minimizing the variational upper bound on the negative log-likelihood of the data $\textbf{x}_0$ under $p_\theta(\textbf{x}_{0:T})$. This objective can be succinctly expressed as:
\begin{equation}
\min_{\theta} \mathbb{E}_{t \sim U\{1,T\}, \textbf{x}_0 \sim p(\textbf{x}_0), \textbf{x}_t \sim q(\textbf{x}_t|\textbf{x}_0)} \left[ \frac{1}{2} || \boldsymbol{\epsilon} - \epsilon_\theta(\textbf{x}_t, t)||^2_2 \right]    
\end{equation}
\textbf{Latent Diffusion Model for 3D Point Cloud:}
In this study, we employ the hierarchical latent diffusion model (LION) proposed in \cite{lion} as the backbone for the TTA task. The LION model leverages a VAE network composed of two hierarchical encoders and one decoder. The first encoder, denoted as \(q_{z}(\textbf{z}_0|\textbf{x})\), converts the input point cloud \(\textbf{x}\) into an abstract latent vector \(\textbf{z}_{0}\in \mathbb{R}^{D_z}\), referred to as the shape latent. Subsequently, another encoder, denoted as \(q_{h}(\textbf{h}_0|\textbf{z}_0, \textbf{x})\), maps the input point cloud to a point cloud-structured latent space, represented by \(\textbf{h}_0\in \mathbb{R}^{4\times n}\), referred to as the latent points. Here, the first three dimensions represent the $xyz$-coordinates, and the last dimension represents an additional feature. Additionally, two denoising diffusion networks denoted by $\boldsymbol{\epsilon}_z\left(\textbf{z}_t, t\right)$ and $\boldsymbol{\epsilon}_h\left(\mathbf{h}_t, \textbf{z}_0, t\right)$   are trained to model the shape latent and latent points, respectively. Finally, the decoder denoted by \(p_{d}(\textbf{x}|\textbf{z}_0, \textbf{h}_0)\) takes the shape latent and latent points as inputs and maps them back to the point cloud. This generative framework is trained in two stages. In the first stage, the encoders and the decoder are simultaneously trained to maximize the variational lower bound over the data log-likelihood:
\begin{equation}
    \begin{aligned}
    \mathcal{L}_{\mathrm{ELBO}} &= \mathbb{E}_{ p(\textbf{x}), q_{z}\left(\textbf{z}_0 \mid \textbf{x}\right), q_{h}\left(\mathbf{h}_0 \mid \textbf{x}, \textbf{z}_0\right)} \Big[ \log p_{d}\left(\textbf{x} \mid \textbf{z}_0, \mathbf{h}_0\right) \\
    &\quad - \gamma_{\textbf{z}} D_{\mathrm{KL}}\left(q_{z}\left(\textbf{z}_0 \mid \textbf{x}\right) \mid p\left(\textbf{z}_0\right)\right) \\
    &\quad - \gamma_{\mathbf{h}} D_{\mathrm{KL}}\left(q_{h}\left(\mathbf{h}_0 \mid \textbf{z}_0, \textbf{x}\right) \mid p\left(\mathbf{h}_0\right)\right) \Big]
    \end{aligned}
\end{equation}

Here, $p(\textbf{z}_0)$ and $p(\textbf{h}_0)$ denote the Gaussian priors imposed on shape latent and latent points, respectively.  In the second stage, the two latent diffusion models are trained on the encodings $\textbf{z}_0$ and
$\textbf{h}_0$ sampled from $q_{z}(\textbf{z}_0|\textbf{x})$ and $q_{h}(\textbf{h}_0|\textbf{z}_0, \textbf{x})$, minimizing the following loss functions:
\begin{equation}
\begin{aligned} &\mathcal{L}_{\mathrm{SM}_z}=\mathbb{E}_{t, \boldsymbol{\epsilon}, p(\textbf{x}), q_{z}\left(\textbf{z}_0 \mid \textbf{x}\right)  }\left\|\boldsymbol{\epsilon}-\boldsymbol{\epsilon}_z\left(\textbf{z}_t, t\right)\right\|_2^2,\\
&\mathcal{L}_{\mathrm{SM_h}}=\mathbb{E}_{t,\boldsymbol{\epsilon}, p(\textbf{x}), q_{z}\left(\textbf{z}_0 \mid \textbf{x}\right), q_{h}\left(\mathbf{h}_0 \mid \textbf{x}, \textbf{z}_0\right)}\left\|\boldsymbol{\epsilon}-\boldsymbol{\epsilon}_{h}\left(\mathbf{h}_t, \textbf{z}_0, t\right)\right\|_2^2
\end{aligned}
\end{equation}




\subsection{Model Overview}
\begin{figure*}[!t]
    \centering
    \includegraphics[width=\textwidth]{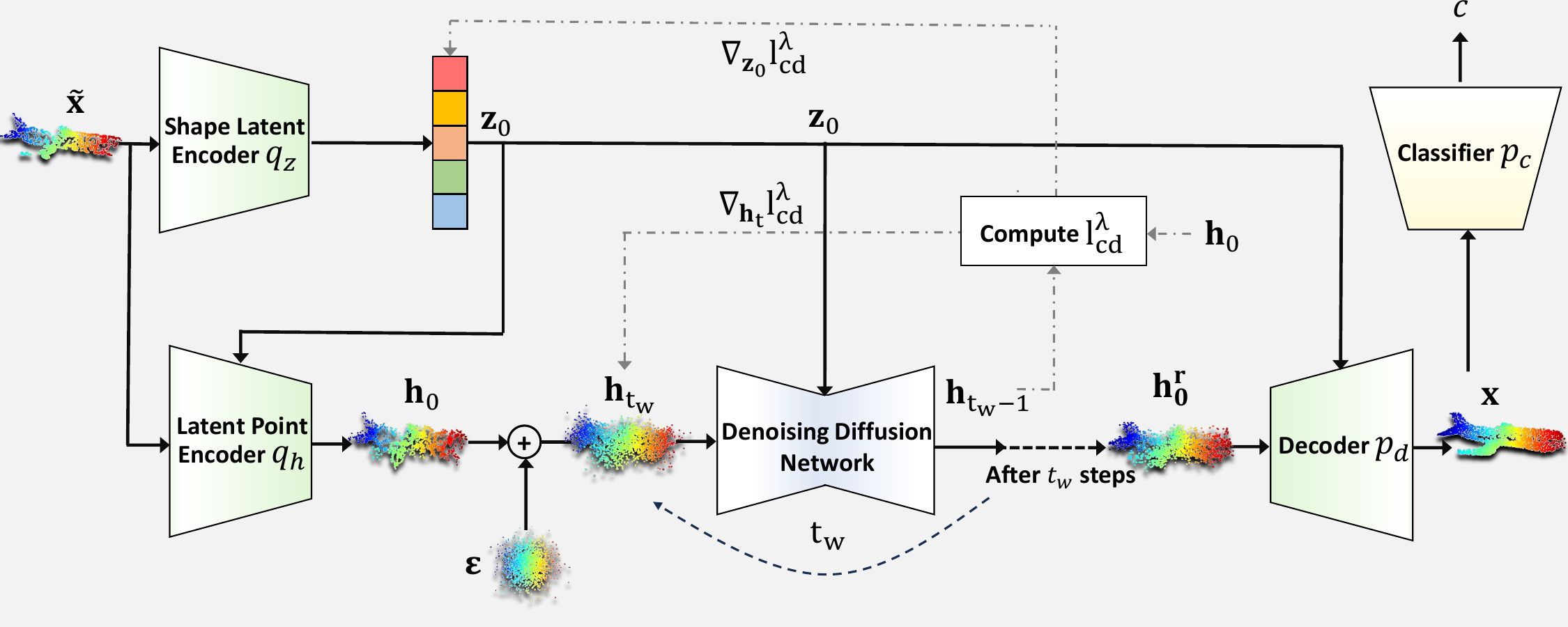}
    \caption{Given a corrupted test point cloud \(\tilde{\textbf{x}}\), we adapt it to the source domain to improve classifier \(p_c\). Using the LION model \cite{lion}, the point cloud is encoded into shape latent \(\textbf{z}_0\) and latent points \(\textbf{h}_0\). The latent points \(\textbf{h}_0\), perturbed with Gaussian noise, are progressively denoised over \(t_w\) iterations. During denoising, both shape latent and latent points are updated to minimize the SCD distance\(l^{\lambda}_{cd}\). 
    After denoising, the latent points are decoded into the adapted point cloud and passed to the classifier \(p_c\) for prediction.
    }
    \label{fig:overview-diagram}
\end{figure*}

We provide an overview of the proposed diffusion-based TTA method, as illustrated in Figure \ref{fig:overview-diagram}. Starting with a corrupted test point cloud \(\tilde{\textbf{x}}\) from domain \(\mathcal{Q}_t\), our objective is to adapt the data back to the source domain \(\mathcal{Q}_s\) to improve the performance of a classifier  \(p_c\) trained on \(\mathcal{Q}_s\). Leveraging the LION model \cite{lion}, the input point cloud \(\tilde{\textbf{x}}\) is first encoded into a shape latent and latent points, represented as \(\textbf{z}_0\) and \(\textbf{h}_0\), respectively. The latent points \(\textbf{h}_0\) are then perturbed with Gaussian noise \(\epsilon \in \mathcal{N}(\textbf{0}, \textbf{I})\) at the \(t_w\) time step using Eq. \ref{eq:direct-forward-diffusion}. The noisy latent points undergo \(t_w\) iterations of denoising through the diffusion network until they sufficiently revert to resemble the source domain point cloud. The denoising process is guided by minimizing our proposed Selective Chamfer Distance (SCD), between the predicted and the original latent points to enhance fidelity, which is denoted by \(l^{\lambda}_{cd}\), where \(\lambda\) is a hyperparameter controlling the strictness of the similarity enforcement. In addition, the shape latent \(\textbf{z}_0\) is updated in the direction of the SCD gradient to improve the conditional signal it provides to the diffusion model during the denoising process. Finally, after $t_w$ steps of denoising, the rectified latent points $\textbf{h}_0^r$ are decoded to generate the adapted point cloud.


\subsection{Denoising Diffusion-based Adaption Method} 


Initially, the corrupted point cloud $\tilde{\textbf{x}}$ is decoded into the shape latent ${\textbf{z}}_0$ and latent points ${\textbf{h}}_0$ using the corresponding encoders, $q_z(.)$ and $q_h(.)$, respectively:
\begin{equation}
    {\textbf{z}}_0 \sim q_{z}\left(\textbf{z}_0 \mid \tilde{\textbf{x}}\right), \quad {\textbf{h}}_0 \sim q_{h}\left(\textbf{h}_0 \mid {\textbf{z}}_0, \, \tilde{\textbf{x}} \right)
\end{equation}
To achieve effective test-time adaptation, the model must remain agnostic to various distribution shifts. To this end, we perturb the latent points with Gaussian noise through the $t_w \in (0,T)$ forward steps to neutralize the effects of different corruptions: \(   {\textbf{h}}_{t_w} = \sqrt{\bar{\alpha}_{t_w}}{\textbf{h}}_{0} + \sqrt{1 - \bar{\alpha}_{t_w}}\boldsymbol{\epsilon}
         \label{eq:perturb_latent_point}\),
where ${\textbf{h}}_{t_w}$ denotes the perturbed latent points and \( \boldsymbol{\epsilon }\) is drawn from a standard Gaussian distribution. Following this, the added noise is estimated using the denoising network $\boldsymbol{\epsilon}_h({\textbf{h}}_{t_w}, \textbf{z}_0, t)$. Subsequently, the latent points of the previous time step ($\textbf{h}_{t_w-1}$) are estimated using the DDIM \cite{ddim} sampling technique, guided by a regularization term denoted as $\mathcal{R}$:

\begin{equation}
{\textbf{h}}_{t_w-1} = \sqrt{\bar{\alpha}_{t_w-1}} \bar{\textbf{h}}_0 + \sqrt{1-\bar{\alpha}_{t_w-1}} \boldsymbol{\epsilon}_h(\textbf{h}_{t_w}, \textbf{z}_0, t_w) + \eta \mathcal{R}
\label{eq:latent_point_updating}
\end{equation}
The first two terms in Eq. \ref{eq:latent_point_updating} constitute the non-markovian deterministic process in DDIM. The regularization term aims to align the predicted latent points to the original ones.  $\bar{\textbf{h}}_0$ denotes the estimated original latent points using the following equation:
\begin{equation}
    \bar{\textbf{h}}_0 = \frac{{\textbf{h}}_{t}-\sqrt{1-\bar{\alpha}_{t}}\boldsymbol{\epsilon}_h(\textbf{h}_{t_w}, \textbf{z}_0, t)}{\sqrt{\bar{\alpha}_t}}
\end{equation}
We introduce and employ the gradient of the Selective Chamfer distance (SCD) denoted as \(l^{\lambda}_{cd}\), with respect to \(\textbf{h}_{t_{w}-1}\) as the regularization term: \( \mathcal{R}=-\nabla_{\textbf{h}_{t_{w}-1}}l^{\lambda}_{cd}\).

Employing this regularization term along with the deterministic denoising process ensures that the latent points are refined towards a coarse representation of original latent points. Since the original latent points are corrupted, we avoid perfect alignment by considering  a fraction of latent points in the SCD distance (\( l^{\lambda}_{cd} \)). To compute \( l^{\lambda}_{cd} \), we sum the lowest \(\lambda\%\) of the sorted minimum squared distances between the two point sets, reducing the influence of outliers and focusing on coarse similarities.
:
\begin{equation}
\begin{aligned}
l^{\lambda}_{cd}\:(\textbf{h}_0, \bar{\textbf{h}}_0) &= \frac{1}{|\textbf{h}_0|} \sum \text{lower}_{\lambda} (\text{sort}(D_{\textbf{h}_0})) \\
&\quad + \frac{1}{|\bar{\textbf{h}}_0|} \sum \text{lower}_{\lambda} (\text{sort}(D_{\bar{\textbf{h}}_0}))
\end{aligned}
\end{equation}

where
$D_{\textbf{h}_0} = \left\{ \min_{\bar{h} \in \bar{\textbf{h}}_0} \|h - \bar{h}\|_2^2 : \forall h \in \textbf{h}_0 \right\}$ and $D_{\bar{\textbf{h}}_0} = \left\{ \min_{h \in \textbf{h}_0} \|\bar{h} - h\|_2^2 : \forall \bar{h} \in \bar{\textbf{h}}_0 \right\}$.

 Additionally, given that the initial shape latent $\textbf{z}_0$, obtained from the input point cloud, potentially leads to inaccurate guidance for the denoising network, we propose an adjustment over the shape latent using the gradient descent technique.  The update formula is expressed as:
\begin{equation}
\textbf{z}_0 \leftarrow \textbf{z}_0 - \gamma \nabla{\textbf{z}_0} l^{\lambda}_{cd}
\label{eq:shape_latent_updating}
\end{equation}

\begin{algorithm}[!t]
\small
\caption{The algorithm for the proposed 3DD-TTA.}
\begin{algorithmic}[1]
 \State \textbf{Input:} Corrupted point cloud $\tilde{\textbf{x}}$, shape encoder $q_z(.)$, latent point encoder $q_h(.)$, decoder $p_d(.)$, diffusion prior $\boldsymbol{\epsilon}_h(.)$, and source classifier $p_c(.)$
\State ${\textbf{z}}_0 \sim q_{z}\left(\textbf{z}_0 \mid \tilde{\textbf{x}}\right)$
 \Comment{obtain the shape latent} 
\State ${\textbf{h}}_0 \sim q_{h}\left(\textbf{h}_0 \mid {\textbf{z}}_0, \tilde{\textbf{x}}\right)$
 \Comment{obtain the latent points}
\State ${\textbf{h}}_{t_w} \gets \sqrt{\bar{\alpha}_{t_w}} \, {\textbf{h}}_{0} + \sqrt{1-\bar{\alpha}_{t_w}} \boldsymbol{\epsilon} $ \Comment{perturb the latent points}
\State \textbf{for} $t \gets t_w ... 1$ \textbf{do}:
\State $\, \quad \boldsymbol{\epsilon}_t \gets \boldsymbol{\epsilon}_h ({\textbf{h}}_{t}, \textbf{z}_0, t)$ \Comment{estimate the diffusion noise by $\boldsymbol{\epsilon}_h(.)$}
    \State \, \quad $\bar{\textbf{h}}_0 \gets \frac{{\textbf{h}}_{t}-\sqrt{1-\bar{\alpha}_{t}}\boldsymbol{\epsilon}_t}{\sqrt{\bar{\alpha}_t}}$ \Comment{estimate original latent points}
\State \, \quad $l^{\lambda}_{cd}=\operatorname{CD}^{\lambda}(\bar{\textbf{h}}_0, {\textbf{h}}_0)$ \Comment{compute SCD distance}
    \State $\, \quad {\textbf{z}}_0 \gets {\textbf{z}}_0 - \gamma \nabla_{{\textbf{z}}_0} l^{\lambda}_{cd}$ \Comment{update the shape latent} 
\State  $\, \quad {\textbf{h}}_{t-1} \gets \sqrt{\bar{\alpha}_{t-1}} \bar{\textbf{h}}_0 + \sqrt{1-\bar{\alpha}_{t-1}} \boldsymbol{\epsilon}_t-\eta \nabla_{{\textbf{h}}_t}l^{\lambda}_{cd}$  \Comment{obtain the latent points for previous time step}
\State $\textbf{x} \gets p_d(\textbf{x}|{\textbf{z}}_0, {\textbf{h}}_0)$ \Comment{reconstruct the point cloud}
\State ${c} \sim p_c({c}|\textbf{x})$ \Comment{pass the point cloud to the source classifier}
\end{algorithmic}
\label{algorithm1}
\end{algorithm}

This gradient descent-based updating technique ensures that the shape latent is adjusted in a direction that enhances the denoising process, resulting in improved alignment between the denoised and the original latent points.
 
We iteratively apply the denoising steps following the update rules in Eq. \ref{eq:latent_point_updating} and \ref{eq:shape_latent_updating}. This process is repeated up to $t_w$ times to approximate $\textbf{h}_0$ and $\textbf{z}_0$.
Finally, we decode the denoised latent points using the decoder: ${\textbf{x}} \sim p_{d}\left({\textbf{x}} \mid {\textbf{z}}_0, {\mathbf{h}}_0\right)$. The reconstructed point cloud $\textbf{x}$ is then fed into the source classifier $p_c$ to determine the final class label: $c \sim p_c(\textbf{x})$.

\begin{table*}[!t]
\centering
\footnotesize
\setlength{\tabcolsep}{1.9pt}
\caption{Classification accuracies on  ShapeNet-c. Point-MAE \cite{point-mae}, as trained in \cite{mirza2023mate}, serves as the source classifier denoted as src. The highest accuracy is in bold, while the second-best is underlined.}
\scalebox{1.3}{
\begin{tabular}{lcccccccccccccccc}
\multicolumn{1}{l}{Methods } & uni & gauss & back & impu & ups & rbf & rbf-i & den-d & den-i & shear & rot & cut & dist & occ &lidar & Mean \\
\hline 
Point-MAE (src) \cite{point-mae} & 72.5 & 66.4 & 15.0 & 60.6 & 72.8 & 72.6 & 73.4 & 85.2 & 85.8 & 74.1 & 42.8 & \underline{84.3} & 71.7 & 8.4 & 4.3 & 59.3 \\
DUA \cite{dua} & 76.1 & 70.1 & 14.3 & 60.9 & 76.2 & 71.6 & 72.9 & 80.0 & 83.8 & {77.1} & $\textbf{57.5}$ & 75.0 & 72.1 & 11.9 & 12.1 & 60.8 \\
TTT-Rot \cite{ttt-rot} & 74.6 & 72.4 & \underline{23.1} & 59.9 & 74.9 & 73.8 & \underline{75.0} & 81.4 & 82.0 & 69.2 & 49.1 & 79.9 & 72.7 & $\underline{14.0}$ & 12.0 & 60.9 \\
SHOT \cite{shot} & 44.8 & 42.5 & 12.1 & 37.6 & 45.0 & 43.7 & 44.2 & 48.4 & 49.4 & 45.0 & 32.6 & 46.3 & 39.1 & 6.2 & 5.9 & 36.2 \\
T3A \cite{t3a} & 70.0 & 60.5 & 6.5 & 40.7 & 67.8 & 67.2 & 68.5 & 79.5 & 79.9 & 72.7 & 42.9 & 79.1 & 66.8 & 7.7 & 5.6 & 54.4 \\
TENT \cite{tent} & 44.5 & 42.9 & 12.4 & 38.0 & 44.6 & 43.3 & 44.3 & 48.7 & 49.4 & 45.7 & 34.8 & 48.6 & 43.0 & 10.0 & 10.9 & 37.4 \\
 MATE-S \cite{mirza2023mate} & \underline{77.8} & \underline{74.7} & 4.3 & \underline{66.2} & \underline{78.6} & \underline{76.3} & \textbf{75.3} & \underline{86.1} & \underline{86.6} & \textbf{79.2} & \underline{56.1} & 84.1 & \underline{76.1} & 12.3 & \underline{13.1} & \underline{63.1} \\
\hline
3DD-TTA (ours) & \textbf{81.6} & \textbf{80.7} & \textbf{77.6} & \textbf{77.2} & \textbf{85.4} & \textbf{76.5} & \textbf{75.3} & \textbf{86.5} & \textbf{88.2} & \underline{77.2}& {50.4} & \textbf{85.4} & \textbf{76.5} & \textbf{14.9} & \textbf{14.2} & $\textbf{69.8}$\\
\hline
\end{tabular}}
\label{table1-shapenet}
\end{table*}
The proposed 3DD-TTA method is summarized in Algorithm~\ref{algorithm1}. Extensive experiments show that our adaptation method effectively restores corrupted point clouds to the source domain while remaining agnostic to corruption types, significantly improving the performance of the source classifier.

\section{Experiments}

\noindent\textbf{Setup:} We adopted the hierarchical latent diffusion model proposed in \cite{lion}, pre-trained on the ShapeNet dataset \cite{shapenet} for our 3DD-TTA model. We employed the deterministic DDIM \cite{ddim} process over 100 total time steps to accelerate the denoising process. Our investigation reveals that perturbing the latent points with noise corresponding to the 5th time step and denoising through these five steps is sufficient to mitigate most of the corruption. For the weights used in updating the shape latent ($\gamma$) and latent points ($\eta$), we conducted a grid search and determined that 0.01 is the optimal setting for both, after testing various configurations. Regarding the SCD distance parameter ($\lambda$), we analyzed different values and found that $\lambda = 0.96$ yields the best performance (refer to Section B in the appendix for more details). All experiments were conducted on a single NVIDIA A6000 GPU.

\subsection{Datasets and Corruption Methods}


\textbf{ShapeNet-c:}  ShapeNet \cite{shapenet}, a large-scale point cloud dataset with 51,127 shapes across 55 categories, is used to train the LION model \cite{lion}, the backbone of our 3DD-TTA framework. Thus, we primarily evaluate our TTA model's performance on this dataset. For experiments, we applied 15 types of corruptions to the test set using the open-source ModelNet40-c implementation \cite{modelnet40-c}, creating a corrupted version called ShapeNet-c. Further details about the corruption function can be found in Appendix A.  

\textbf{ModelNet40-c:} ModelNet40-c \cite{modelnet40-c} introduces the aforementioned 15 typical corruptions to the original test set of ModelNet40 \cite{modelnet40}. We use this dataset to test the generalization power of the proposed 3DD-TTA approach on datasets beyond the source datasets.

\textbf{ScanObjectNN-c:} ScanObjectNN \cite{scanobjectnn}, a real-world point cloud dataset with 15 categories, is corrupted using the same open-source code as ModelNet40-c \cite{modelnet40-c}, introducing 15 corruptions into the test set. We refer to this dataset as ScanObjectNN-c.

\subsection{Baselines}

We compared our 3DD-TTA model to the non-adapted source classifier, PointMAE (src) \cite{point-mae}, and six TTA approaches originally proposed for images, each assuming access to a single sample for test-time adaptation. The baselines include: (1) SHOT \cite{shot}, which minimizes output entropy; (2) T3A \cite{t3a}, which learns class-specific prototypes to replace the pre-trained classifier; (3) TENT \cite{tent}, which also minimizes prediction entropy; (4) DUA \cite{dua}, which updates batch normalization statistics to adapt at test time; (5) TTT-Rot \cite{ttt-rot}, which adapts via self-supervised rotation prediction tasks; and (6) MATE-S \cite{mirza2023mate}, which fine-tunes PointMAE \cite{point-mae} for reconstruction at test time.

\subsection{Results}

\begin{figure*}[!h]
    \centering
    \includegraphics[width=0.9\textwidth]{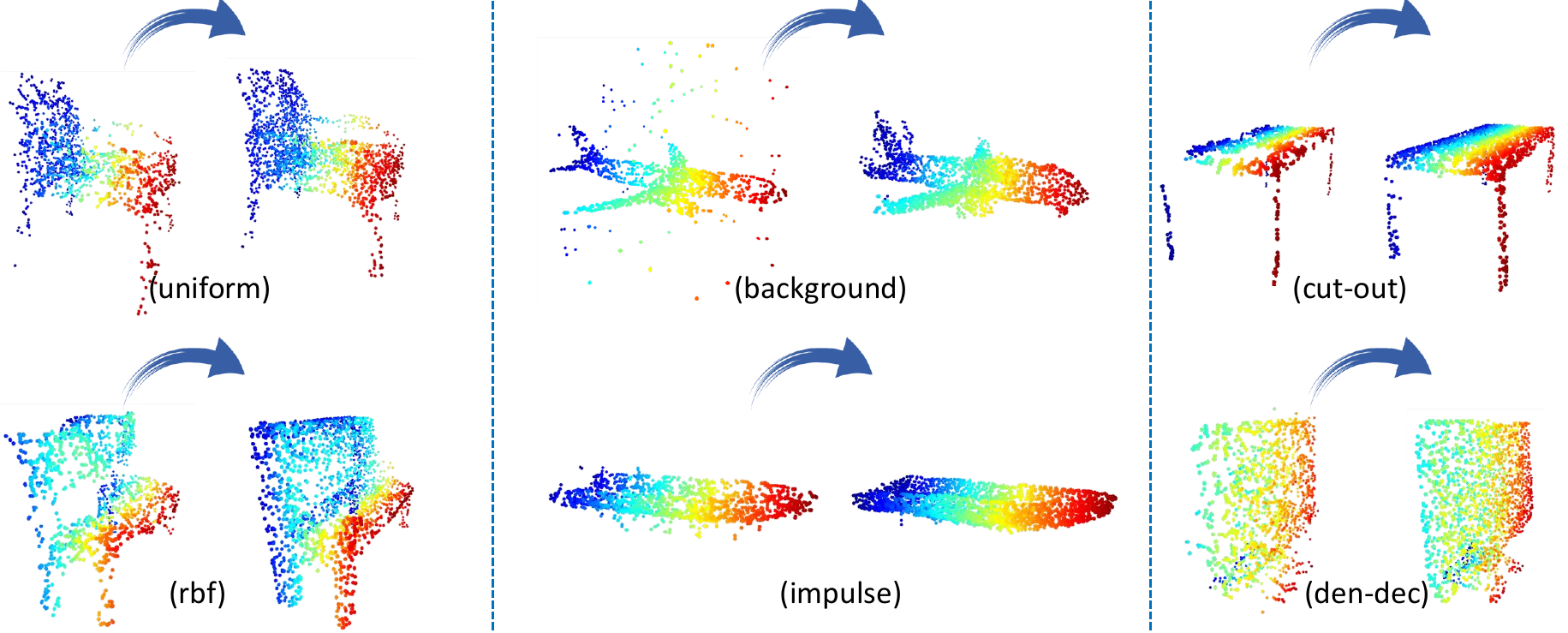}
    \caption{Qualitative assessment of the proposed test-time adaptation across various corruptions.}
    \label{fig:samples}
\end{figure*}


\begin{table*}[!t]
\caption{Classification accuracies on ModelNet40-c. Point-MAE \cite{point-mae}, as trained in \cite{mirza2023mate}, serves as the source classifier denoted as src. The highest accuracy is in bold, while the second-best is underlined.}
\centering
\footnotesize
\setlength{\tabcolsep}{1.9pt}
\scalebox{1.3}{
\begin{tabular}{lcccccccccccccccc}
\multicolumn{1}{l}{Methods } & uni & gauss & back & impu & ups & rbf & rbf-i & den-d & den-i & shear & rot & cut & dist & occ &lidar & Mean \\
\hline

Point-MAE (src) \cite{point-mae} & 62.4 & 57.0 & 32.0 & 58.8 & 72.1 & 61.4 & 64.2 & 75.1 & 80.8 & 67.6 & 31.3 & 70.4 & 64.8 & 36.2 & 29.1 & 57.6 \\
DUA \cite{dua}& 65.0 & 58.5 & 14.7 & 48.5 & 68.8 & 62.8 & 63.2 & 62.1 & 66.2 & {68.8} & $\textbf{{46.2}}$ & 53.8 & 64.7 & $\textbf{{41.2}}$ & $\underline{36.5}$ & 54.7 \\
TTT-Rot \cite{ttt-rot} & 61.3 & 58.3 & \underline{{34.5}} & 48.9 & 66.7 & 63.6 & 63.9 & 59.8 & 68.6 & 55.2 & 27.3 & 54.6 & 64.0 & 40.0 & 29.1 & 53.0 \\
SHOT \cite{shot} & 29.6 & 28.2 & 9.8 & 25.4 & 32.7 & 30.3 & 30.1 & 30.9 & 31.2 & 32.1 & 22.8 & 27.3 & 29.4 & 20.8 & 18.6 & 26.6 \\
T3A & 64.1 & 62.3 & 33.4 & 65.0 & 75.4 & 63.2 & 66.7 & 57.4 & 63.0 & \underline{72.7} & 32.8 & 54.4 & {67.7} & 39.1 & 18.3 & 55.7 \\
TENT \cite{tent} & 29.2 & 28.7 & 10.1 & 25.1 & 33.1 & 30.3 & 29.1 & 30.4 & 31.5 & 31.8 & 22.7 & 27.0 & 28.6 & 20.7 & 19.0 & 26.5 \\
 MATE-S \cite{mirza2023mate} & $\underline{{75.0}}$ & \underline{71.1} & 27.5 & $\underline{67.5}$ & $\underline{78.7}$ & $\textbf{{69.5}}$ & $\textbf{{72.0}}$ & $\underline{79.1}$ & \underline{84.5} & $\textbf{{75.4}}$ & \underline{44.4} & \underline{73.6} & $\textbf{{72.9}}$ & 39.7 & 34.2 & $\underline{64.3}$ \\
\hline
3DD-TTA (ours) & \textbf{77.5} & \textbf{79.1} & \textbf{49.9} & \textbf{80.3} & \textbf{81.8} & \underline{63.8}& \underline{66.9} &\textbf{{79.3}}&\textbf{{84.7}}&63.7&33.4&\textbf{{74.7}}&\underline{68.2}&\underline{39.9}& \textbf{{42.2}}& \textbf{{65.7}}\\
\hline
\end{tabular}}
\label{table 2-modelnet40-c}
\end{table*}
Table \ref{table1-shapenet} compares the performance of various TTA approaches on the ShapeNet-c dataset for different types of corruptions. The table clearly shows that the 3DD-TTA method outperforms the other methods by a significant margin in most corruption types. In particular, 3DD-TTA excels at mitigating noise-related corruptions such as uniform, Gaussian, impulse, and background noise. This effectiveness is attributed to the denoising capabilities of DDMs during the reverse diffusion process. Notably, 3DD-TTA dramatically boosts the source classifier's performance on background noise, raising accuracy from 15.0$\%$ to 77.6$\%$. 
 In addition, our 3DD-TTA  outperforms other TTA frameworks on density-based corruptions such as cut-out and density increase. However, the model faces limitations in addressing the transformation-based deformations like shear and rotation. This limitation is due to the training-free nature of the model, making it challenging to reverse transformations to their original shape without additional training. Figure \ref{fig:samples} shows the qualitative performance of the 3DD-TTA algorithm in reconstructing the corrupted point clouds. The model successfully restores the point clouds, thereby improving the performance of the source classifier. More qualitative results are provided in Appendix D.

\vspace{4mm}
We suggest that our model, built on the LION backbone \cite{lion}, which was pre-trained on the extensive ShapeNet dataset \cite{shapenet}, can effectively generalize to reconstruct corrupted point clouds from other datasets. To validate this, we conduct experiments on the ModelNet40-c and ScanObjectNN-c datasets. Table \ref{table 2-modelnet40-c} presents the performance of various TTA models in addressing different corruptions within the ModelNet40-c dataset \cite{modelnet40-c}. Notably, the diffusion model is effective in addressing the noise-related corruptions in ModelNet40-c dataset \cite{modelnet40-c}, despite being trained on the ShapeNet dataset \cite{shapenet}.  Similarly, the model outperforms other methods in addressing density-related corruptions but is less effective for transformation-based corruptions, ranking second or third for these deformations.
We also conducted experiments on the corrupted version of the real-world ScanObjectNN dataset \cite{scanobjectnn}, which inherently suffers from noise, background issues, and occlusion. Table \ref{tab:table3} displays the performance of the 3DD-TTA model on this dataset, demonstrating its strong generalization capability to real-world datasets. 3DD-TTA boosts Point-MAE (source) in most corruption types, showing improved robustness across different corruptions.

\begin{table*}
\caption{Classification accuracies on ScanObjectNN-c dataset.}
\centering
\footnotesize
\setlength{\tabcolsep}{1.9pt}
\scalebox{1.3}{
\begin{tabular}{lcccccccccccccccc}
\multicolumn{1}{l}{Corruptions: } & uni & gauss & back & impu & ups & rbf & rbf-i & den-d & den-i & shear & rot & cut & dist & occ &lidar & Mean \\
\hline 
Point-MAE (src) \cite{point-mae} & { {59.0}} & { {59.0}} & {57.3} & { 66.6} & {  64.4} & {  \textbf{73.6}} & {  72.9} & {  {\textbf{81.2}}} & {  82.4} & {  \textbf{71.9}} & {  59.7} & {  78.3} & {  70.7}  & {  14.1}& {  11.0}& {  61.5} \\

\hline

3DD-TTA (ours)   & {  \textbf{68.7}}  & {  \textbf{68.0} }  &  {  {\textbf{61.1}}} & {  {\textbf{67.3}}}  & {  {\textbf{74.0}}} & {  {73.0}} & {  \textbf{73.8}} & {  {78.8}}& {  {\textbf{84.7}}}  & {  {70.9}} &{  {\textbf{59.8}}} &  {  {\textbf{81.4}}} & {  \textbf{72.3}} & {  \textbf{14.8}}& {  \textbf{{16.5}}}& {  \textbf{64.3}} \\
\hline
\end{tabular}}
\label{tab:table3}

\end{table*}

\subsection{Ablation Study}
\label{section:ablation}
 
\begin{figure*}[!t]
    \centering
    \includegraphics[width=\textwidth]{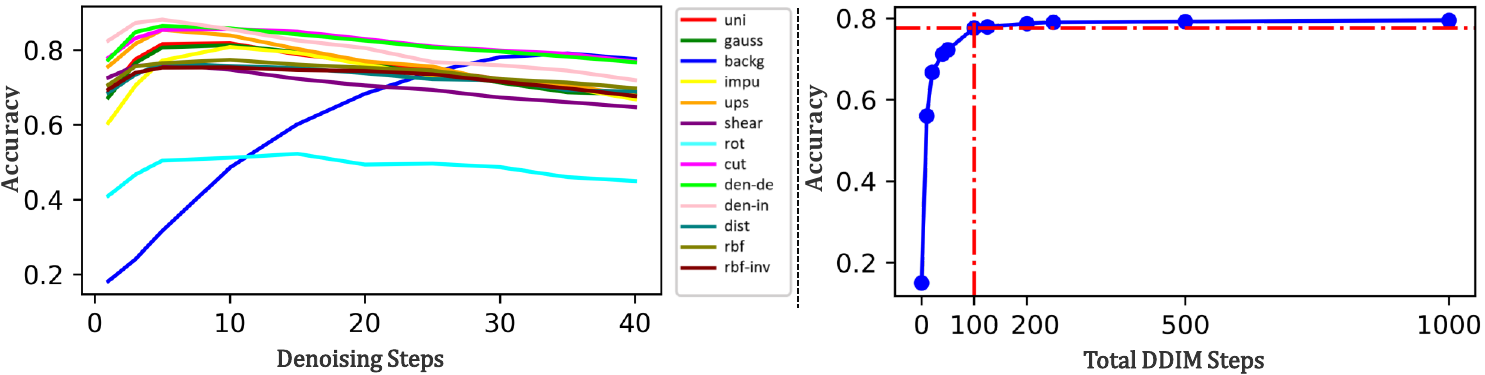}
    \caption{ (\textbf{left}) Accuracy of the source classifier after adaptation using different numbers of denoising steps. (\textbf{right}) Performance of the model across different numbers of total DDIM steps.}
    \label{fig:ablation-steps}
    \vspace{-5mm}
\end{figure*}

\noindent\textbf{Number of Denoising Steps for Reconstruction:}
While the denoising diffusion network in the original LION \cite{lion} model has been trained with 1000 time steps, we posit that a total of 100 deterministic DDIM time steps is sufficient for sampling a point cloud from pure Gaussian noise without compromising point cloud quality. This implies a 10-step skip for each DDIM step. 
We base our assumption on DDIMs' efficiency in achieving high-quality generation with fewer steps and the lack of high-frequency details in point cloud meshes, meaning that reducing denoising steps should not significantly impact the quality of the generated point clouds. As the first experiment, we fixed the total number of DDIM steps to 100 and investigated the optimal number of denoising steps for reconstructing corrupted point clouds. Our analysis, illustrated in Figure \ref{fig:ablation-steps} (left), demonstrates that five denoising steps are sufficient for most types of corruption. However, for extreme outliers such as background noise, a more extensive denoising process of up to 35 steps is required. This is because outliers and isolated points in the point cloud require significant perturbation to fill the gaps with noise, followed by denoising to return the point cloud to its natural manifold.

\noindent\textbf{Adaptation Time vs. Number of Denoising Steps:} Table \ref{tab:table4} shows the time duration required for test-time adaptation using our method as the number of denoising steps increases. The adaptation time grows linearly with the number of denoising steps, ranging from 12 ms for 1 step to 290 ms for 40 steps. As only 5 denoising steps (taking 40 ms) are sufficient for most corruptions, the proposed method is efficient, making it suitable for time-sensitive applications.
\begin{table}[!h]
\caption{Adaptation time (ms) for different numbers of denoising steps in the 3DD-TTA method.}
\centering
\footnotesize
\setlength{\tabcolsep}{1.9pt}
\scalebox{1.3}{
\begin{tabular}{lccccccc}
\multicolumn{1}{l}{$\#$ Denoising Steps: } & & 1 & 5 & 10 & 20 & 30 & 40 \\
\hline 
Time Duration (ms) & & 12 & 40 & 74 & 142 & 218 & 290 \\
\hline
\end{tabular}}
\label{tab:table4}
\end{table}

\noindent\textbf{Number of Total DDIM Steps:}
In a second experiment, we investigated the impact of the total number of DDIM steps to validate our assumption of 100 steps. We chose the background noise for this analysis because it requires more denoising steps to resolve, thereby better illustrating the impact of reducing DDIM steps. We fixed 35 denoising steps for reconstruction with 100 total DDIM steps, then varied the total DDIM steps (S) from 0 to 1000, proportionally adjusting the denoising steps to 35*(S/100) to assess its impact on source classifier performance. Figure \ref{fig:ablation-steps} (right) shows that accuracy remains stable with fewer DDIM steps until around 100 steps, after which a sharp decline occurs. Based on this, we selected 100 DDIM steps to balance adaptation time and accuracy.

\noindent\textbf{Limitation:} Our model performs exceptionally well with just five denoising steps for most types of corruption, making it efficient and suitable for time-sensitive applications. However, severe outliers, such as background noise, may require up to 35 steps, increasing the processing time for these cases. Additionally, future work could explore determining the optimal number of denoising steps adaptively based on the input point cloud. In addition, while the model effectively addresses density inconsistencies and noise-based corruptions, it struggles with transformation-based deformations like rotation and shear. Making parts of the network trainable could help handle both types of corruption in the future. 


\section{Conclusion}

This paper introduces a novel approach to 3D point cloud test-time adaptation, addressing the challenge of managing discrepancies between training and testing samples, especially in corrupted point clouds. Unlike existing methods that often fine-tune pre-trained models using self-supervised learning, the proposed method focuses on adapting input point cloud samples to the source domain through a denoising diffusion strategy. By leveraging a diffusion model trained on the source domain, the approach iteratively denoises the perturbed test point clouds, enhancing their consistency with the source domain without adjusting the model parameters. Incorporating the proposed updating strategy based on the gradient of the Selective Chamfer Distance (SCD) ensures the generation of high-fidelity, noise-free test samples. Extensive experiments conducted on ShapeNet \cite{shapenet}, ModelNet40 \cite{modelnet40}, and ScanObjectNN \cite{scanobjectnn} datasets validate the efficacy of the proposed approach, demonstrating new state-of-the-art results. This innovative method not only advances the field of 3D test-time adaptation but also provides a promising direction for handling real-world challenges in various domains.



{\small
\bibliographystyle{ieee_fullname}
\bibliography{egbib}

\begin{thebibliography}{10}\itemsep=-1pt

\bibitem{ahmadi2024accv}
Sahar Ahmadi, Ali Cheraghian, Morteza Saberi, Md.~Towsif Abir, Hamidreza Dastmalchi, Farookh Hussain, and Shafin Rahman.
\newblock Foundation model-powered 3d few-shot class incremental learning via training-free adaptor.
\newblock In {\em 16th Asian Conference on Computer Vision (ACCV) 2024}. ACCV, 2024.

\bibitem{3d-diffusion-gradient}
Ruojin Cai, Guandao Yang, Hadar Averbuch-Elor, Zekun Hao, Serge Belongie, Noah Snavely, and Bharath Hariharan.
\newblock Learning gradient fields for shape generation.
\newblock In {\em Computer Vision--ECCV 2020: 16th European Conference, Glasgow, UK, August 23--28, 2020, Proceedings, Part III 16}, pages 364--381. Springer, 2020.

\bibitem{shapenet}
Angel~X Chang, Thomas Funkhouser, Leonidas Guibas, Pat Hanrahan, Qixing Huang, Zimo Li, Silvio Savarese, Manolis Savva, Shuran Song, Hao Su, et~al.
\newblock Shapenet: An information-rich 3d model repository.
\newblock {\em arXiv preprint arXiv:1512.03012}, 2015.

\bibitem{C3PR_ECCV}
Ali Cheraghian, Zeeshan Hayder, Sameera Ramasinghe, Shafin Rahman, Javad Jafaryahya, Lars Petersson, and Mehrtash Harandi.
\newblock Canonical shape projection is all you need for 3d few-shot class incremental learning.
\newblock In {\em European Conference on Computer Vision (ECCV)}. Springer, 2024.

\bibitem{chowdhury2022few}
Townim Chowdhury, Ali Cheraghian, Sameera Ramasinghe, Sahar Ahmadi, Morteza Saberi, and Shafin Rahman.
\newblock Few-shot class-incremental learning for 3d point cloud objects.
\newblock In Shai Avidan, Gabriel Brostow, Moustapha Ciss{\'e}, Giovanni~Maria Farinella, and Tal Hassner, editors, {\em Computer Vision -- ECCV 2022}, pages 204--220, Cham, 2022. Springer Nature Switzerland.

\bibitem{dobler2022robust}
Mario D{\"o}bler, Robert~A Marsden, and Bin Yang.
\newblock Robust mean teacher for continual and gradual test-time adaptation.
\newblock {\em arXiv preprint arXiv:2211.13081}, 2022.

\bibitem{fan2022self}
Hehe Fan, Xiaojun Chang, Wanyue Zhang, Yi Cheng, Ying Sun, and Mohan Kankanhalli.
\newblock Self-supervised global-local structure modeling for point cloud domain adaptation with reliable voted pseudo labels.
\newblock In {\em Proceedings of the IEEE/CVF Conference on Computer Vision and Pattern Recognition}, pages 6377--6386, 2022.

\bibitem{rbf}
Davide Forti and Gianluigi Rozza.
\newblock Efficient geometrical parametrisation techniques of interfaces for reduced-order modelling: application to fluid--structure interaction coupling problems.
\newblock {\em International Journal of Computational Fluid Dynamics}, 28(3-4):158--169, 2014.

\bibitem{gao2023back}
Jin Gao, Jialing Zhang, Xihui Liu, Trevor Darrell, Evan Shelhamer, and Dequan Wang.
\newblock Back to the source: Diffusion-driven adaptation to test-time corruption.
\newblock In {\em Proceedings of the IEEE/CVF Conference on Computer Vision and Pattern Recognition}, pages 11786--11796, 2023.

\bibitem{hatem2023point}
Ahmed Hatem, Yiming Qian, and Yang Wang.
\newblock Point-tta: Test-time adaptation for point cloud registration using multitask meta-auxiliary learning.
\newblock In {\em Proceedings of the IEEE/CVF International Conference on Computer Vision}, pages 16494--16504, 2023.

\bibitem{hatem2023test}
Ahmed Hatem, Yiming Qian, and Yang Wang.
\newblock Test-time adaptation for point cloud upsampling using meta-learning.
\newblock {\em arXiv preprint arXiv:2308.16484}, 2023.

\bibitem{ddpm}
Jonathan Ho, Ajay Jain, and Pieter Abbeel.
\newblock Denoising diffusion probabilistic models.
\newblock {\em Advances in neural information processing systems}, 33:6840--6851, 2020.

\bibitem{hong2023pointcam}
Jie Hong, Shi Qiu, Weihao Li, Saeed Anwar, Mehrtash Harandi, Nick Barnes, and Lars Petersson.
\newblock Pointcam: Cut-and-mix for open-set point cloud learning.
\newblock {\em arXiv preprint arXiv:2212.02011}, 2023.

\bibitem{t3a}
Yusuke Iwasawa and Yutaka Matsuo.
\newblock Test-time classifier adjustment module for model-agnostic domain generalization.
\newblock {\em Advances in Neural Information Processing Systems}, 34:2427--2440, 2021.

\bibitem{pointcnn2018}
Yangyan Li, Rui Bu, Mingchao Sun, Wei Wu, Xinhan Di, and Baoquan Chen.
\newblock Pointcnn: Convolution on x-transformed points.
\newblock In {\em NeurIPS}, 2018.

\bibitem{liang2020we}
Jian Liang, Dapeng Hu, and Jiashi Feng.
\newblock Do we really need to access the source data? source hypothesis transfer for unsupervised domain adaptation.
\newblock In {\em International conference on machine learning}, pages 6028--6039. PMLR, 2020.

\bibitem{shot}
Jian Liang, Dapeng Hu, and Jiashi Feng.
\newblock Do we really need to access the source data? source hypothesis transfer for unsupervised domain adaptation.
\newblock In {\em International conference on machine learning}, pages 6028--6039. PMLR, 2020.

\bibitem{RS-CNN2019}
Yongcheng Liu, Bin Fan, Shiming Xiang, and Chunhong Pan.
\newblock Relation-shape convolutional neural network for point cloud analysis.
\newblock In {\em CVPR}, 2019.

\bibitem{liu2021ttt++}
Yuejiang Liu, Parth Kothari, Bastien Van~Delft, Baptiste Bellot-Gurlet, Taylor Mordan, and Alexandre Alahi.
\newblock Ttt++: When does self-supervised test-time training fail or thrive?
\newblock {\em Advances in Neural Information Processing Systems}, 34:21808--21820, 2021.

\bibitem{3d-diffusion}
Shitong Luo and Wei Hu.
\newblock Diffusion probabilistic models for 3d point cloud generation.
\newblock In {\em Proceedings of the IEEE/CVF Conference on Computer Vision and Pattern Recognition}, pages 2837--2845, 2021.

\bibitem{dua}
M~Jehanzeb Mirza, Jakub Micorek, Horst Possegger, and Horst Bischof.
\newblock The norm must go on: Dynamic unsupervised domain adaptation by normalization.
\newblock In {\em Proceedings of the IEEE/CVF Conference on Computer Vision and Pattern Recognition}, pages 14765--14775, 2022.

\bibitem{mirza2023mate}
M~Jehanzeb Mirza, Inkyu Shin, Wei Lin, Andreas Schriebl, Kunyang Sun, Jaesung Choe, Mateusz Kozinski, Horst Possegger, In~So Kweon, Kuk-Jin Yoon, et~al.
\newblock Mate: toencoders are online 3d test-time learners.
\newblock In {\em Proceedings of the IEEE/CVF International Conference on Computer Vision}, pages 16709--16718, 2023.

\bibitem{glide}
Alex Nichol, Prafulla Dhariwal, Aditya Ramesh, Pranav Shyam, Pamela Mishkin, Bob McGrew, Ilya Sutskever, and Mark Chen.
\newblock Glide: Towards photorealistic image generation and editing with text-guided diffusion models.
\newblock {\em arXiv preprint arXiv:2112.10741}, 2021.

\bibitem{point-mae}
Yatian Pang, Wenxiao Wang, Francis~EH Tay, Wei Liu, Yonghong Tian, and Li Yuan.
\newblock Masked autoencoders for point cloud self-supervised learning.
\newblock In {\em European conference on computer vision}, pages 604--621. Springer, 2022.

\bibitem{SPHNet2019}
Adrien Poulenard, Marie-Julie Rakotosaona, Yann Ponty, and Maks Ovsjanikov.
\newblock Effective rotation-invariant point cnn with spherical harmonics kernels.
\newblock In {\em 3DV}, 2019.

\bibitem{prabhudesai2023diffusion}
Mihir Prabhudesai, Tsung-Wei Ke, Alexander~Cong Li, Deepak Pathak, and Katerina Fragkiadaki.
\newblock Diffusion-tta: Test-time adaptation of discriminative models via generative feedback.
\newblock In {\em Thirty-seventh Conference on Neural Information Processing Systems}, 2023.

\bibitem{qi2017pointnet}
Charles~R Qi, Hao Su, Kaichun Mo, and Leonidas~J Guibas.
\newblock Pointnet: Deep learning on point sets for 3d classification and segmentation.
\newblock In {\em Proceedings of the IEEE conference on computer vision and pattern recognition}, pages 652--660, 2017.

\bibitem{pointnet2017}
Charles~R Qi, Hao Su, Kaichun Mo, and Leonidas~J Guibas.
\newblock Pointnet: Deep learning on point sets for 3d classification and segmentation.
\newblock In {\em CVPR}, 2017.

\bibitem{pointnet++}
Charles~R. Qi, Li Yi, Hao Su, and Leonidas~J. Guibas.
\newblock Pointnet++: Deep hierarchical feature learning on point sets in a metric space.
\newblock In {\em Proceedings of the 31st International Conference on Neural Information Processing Systems}, NIPS'17, page 5105–5114, Red Hook, NY, USA, 2017. Curran Associates Inc.

\bibitem{pointnet++2017}
Charles~Ruizhongtai Qi, Li Yi, Hao Su, and Leonidas~J Guibas.
\newblock Pointnet++: Deep hierarchical feature learning on point sets in a metric space.
\newblock In {\em NeurIPS}, 2017.

\bibitem{dalle2}
Aditya Ramesh, Prafulla Dhariwal, Alex Nichol, Casey Chu, and Mark Chen.
\newblock Hierarchical text-conditional image generation with clip latents.
\newblock {\em arXiv preprint arXiv:2204.06125}, 1(2):3, 2022.

\bibitem{SFCNN2019}
Yongming Rao, Jiwen Lu, and Jie Zhou.
\newblock Spherical fractal convolutional neural networks for point cloud recognition.
\newblock In {\em CVPR}, 2019.

\bibitem{latent-diffusion-model}
Robin Rombach, Andreas Blattmann, Dominik Lorenz, Patrick Esser, and Bj{\"o}rn Ommer.
\newblock High-resolution image synthesis with latent diffusion models.
\newblock In {\em Proceedings of the IEEE/CVF conference on computer vision and pattern recognition}, pages 10684--10695, 2022.

\bibitem{imagen}
Chitwan Saharia, William Chan, Saurabh Saxena, Lala Li, Jay Whang, Emily~L Denton, Kamyar Ghasemipour, Raphael Gontijo~Lopes, Burcu Karagol~Ayan, Tim Salimans, et~al.
\newblock Photorealistic text-to-image diffusion models with deep language understanding.
\newblock {\em Advances in neural information processing systems}, 35:36479--36494, 2022.

\bibitem{sdxl-turbo}
Axel Sauer, Dominik Lorenz, Andreas Blattmann, and Robin Rombach.
\newblock Adversarial diffusion distillation.
\newblock {\em arXiv preprint arXiv:2311.17042}, 2023.

\bibitem{shin2022mm}
Inkyu Shin, Yi-Hsuan Tsai, Bingbing Zhuang, Samuel Schulter, Buyu Liu, Sparsh Garg, In~So Kweon, and Kuk-Jin Yoon.
\newblock Mm-tta: multi-modal test-time adaptation for 3d semantic segmentation.
\newblock In {\em Proceedings of the IEEE/CVF Conference on Computer Vision and Pattern Recognition}, pages 16928--16937, 2022.

\bibitem{ddm_thermodynamics}
Jascha Sohl-Dickstein, Eric Weiss, Niru Maheswaranathan, and Surya Ganguli.
\newblock Deep unsupervised learning using nonequilibrium thermodynamics.
\newblock In {\em International conference on machine learning}, pages 2256--2265. PMLR, 2015.

\bibitem{ddim}
Jiaming Song, Chenlin Meng, and Stefano Ermon.
\newblock Denoising diffusion implicit models.
\newblock {\em arXiv preprint arXiv:2010.02502}, 2020.

\bibitem{song2023target}
Kaiyu Song and Hanjiang Lai.
\newblock Target to source: Guidance-based diffusion model for test-time adaptation.
\newblock {\em arXiv preprint arXiv:2312.05274}, 2023.

\bibitem{modelnet40-c}
Jiachen Sun, Qingzhao Zhang, Bhavya Kailkhura, Zhiding Yu, Chaowei Xiao, and Z~Morley Mao.
\newblock Benchmarking robustness of 3d point cloud recognition against common corruptions.
\newblock {\em arXiv preprint arXiv:2201.12296}, 2022.

\bibitem{sun2020test}
Yu Sun, Xiaolong Wang, Zhuang Liu, John Miller, Alexei Efros, and Moritz Hardt.
\newblock Test-time training with self-supervision for generalization under distribution shifts.
\newblock In {\em International conference on machine learning}, pages 9229--9248, 2020.

\bibitem{ttt-rot}
Yu Sun, Xiaolong Wang, Zhuang Liu, John Miller, Alexei Efros, and Moritz Hardt.
\newblock Test-time training with self-supervision for generalization under distribution shifts.
\newblock In {\em International conference on machine learning}, pages 9229--9248, 2020.

\bibitem{tsai2024gda}
Yun-Yun Tsai, Fu-Chen Chen, Albert~YC Chen, Junfeng Yang, Che-Chun Su, Min Sun, and Cheng-Hao Kuo.
\newblock Gda: Generalized diffusion for robust test-time adaptation.
\newblock In {\em Proceedings of the IEEE/CVF Conference on Computer Vision and Pattern Recognition}, pages 23242--23251, 2024.

\bibitem{scanobjectnn}
Mikaela~Angelina Uy, Quang-Hieu Pham, Binh-Son Hua, Duc~Thanh Nguyen, and Sai-Kit Yeung.
\newblock Revisiting point cloud classification: A new benchmark dataset and classification model on real-world data.
\newblock In {\em International Conference on Computer Vision (ICCV)}, 2019.

\bibitem{lion}
Arash Vahdat, Francis Williams, Zan Gojcic, Or Litany, Sanja Fidler, Karsten Kreis, et~al.
\newblock Lion: Latent point diffusion models for 3d shape generation.
\newblock {\em Advances in Neural Information Processing Systems}, 35:10021--10039, 2022.

\bibitem{LocalSpecGCN2018}
Chu Wang, Babak Samari, and Kaleem Siddiqi.
\newblock Local spectral graph convolution for point set feature learning.
\newblock In {\em ECCV}, 2018.

\bibitem{tent}
Dequan Wang, Evan Shelhamer, Shaoteng Liu, Bruno Olshausen, and Trevor Darrell.
\newblock Tent: Fully test-time adaptation by entropy minimization.
\newblock In {\em International Conference on Learning Representations}, 2021.

\bibitem{wang2021tent}
Dequan Wang, Evan Shelhamer, Shaoteng Liu, Bruno Olshausen, and Trevor Darrell.
\newblock Tent: Fully test-time adaptation by entropy minimization.
\newblock In {\em International Conference on Learning Representations}, 2021.

\bibitem{Wang_2024_WACV}
Yanshuo Wang, Jie Hong, Ali Cheraghian, Shafin Rahman, David Ahmedt-Aristizabal, Lars Petersson, and Mehrtash Harandi.
\newblock Continual test-time domain adaptation via dynamic sample selection.
\newblock In {\em Proceedings of the IEEE/CVF Winter Conference on Applications of Computer Vision (WACV)}, pages 1701--1710, January 2024.

\bibitem{dgcnn2019}
Yue Wang, Yongbin Sun, Ziwei Liu, Sanjay~E Sarma, Michael~M Bronstein, and Justin~M Solomon.
\newblock Dynamic graph cnn for learning on point clouds.
\newblock {\em Acm Transactions On Graphics (tog)}, 2019.

\bibitem{pointconv2019}
Wenxuan Wu, Zhongang Qi, and Li Fuxin.
\newblock {PointCONV: Deep convolutional networks on 3D point clouds}.
\newblock In {\em CVPR}, 2019.

\bibitem{modelnet40}
Zhirong Wu, Shuran Song, Aditya Khosla, Fisher Yu, Linguang Zhang, Xiaoou Tang, and Jianxiong Xiao.
\newblock 3d shapenets: A deep representation for volumetric shapes.
\newblock In {\em Proceedings of the IEEE conference on computer vision and pattern recognition}, pages 1912--1920, 2015.

\bibitem{xiang2021walk}
Tiange Xiang, Chaoyi Zhang, Yang Song, Jianhui Yu, and Weidong Cai.
\newblock Walk in the cloud: Learning curves for point clouds shape analysis.
\newblock In {\em Proceedings of the IEEE/CVF International Conference on Computer Vision}, pages 915--924, 2021.

\bibitem{spidercnn2018}
Yifan Xu, Tianqi Fan, Mingye Xu, Long Zeng, and Yu Qiao.
\newblock Spidercnn: Deep learning on point sets with parameterized convolutional filters.
\newblock In {\em ECCV}, 2018.

\bibitem{zhang2023starting}
Renrui Zhang, Liuhui Wang, Yali Wang, Peng Gao, Hongsheng Li, and Jianbo Shi.
\newblock Starting from non-parametric networks for 3d point cloud analysis.
\newblock In {\em Proceedings of the IEEE/CVF Conference on Computer Vision and Pattern Recognition}, pages 5344--5353, 2023.

\bibitem{PointGCN2018}
Yingxue Zhang and Michael Rabbat.
\newblock A graph-cnn for 3d point cloud classification.
\newblock In {\em ICASSP}, 2018.

\bibitem{zhao2021point}
Hengshuang Zhao, Li Jiang, Jiaya Jia, Philip~HS Torr, and Vladlen Koltun.
\newblock Point transformer.
\newblock In {\em Proceedings of the IEEE/CVF International Conference on Computer Vision}, pages 16259--16268, 2021.

\bibitem{point-voxel-3d-diffusion}
Linqi Zhou, Yilun Du, and Jiajun Wu.
\newblock 3d shape generation and completion through point-voxel diffusion.
\newblock In {\em Proceedings of the IEEE/CVF international conference on computer vision}, pages 5826--5835, 2021.

\end{thebibliography}
}

\end{document}